\documentclass[preprint,9pt]{elsarticle}
\usepackage{geometry}
\usepackage{amssymb}
\usepackage{amsmath}

\usepackage{amsmath,amsfonts}

\usepackage{algorithmic}
\usepackage{array}
\usepackage[caption=false,font=normalsize,labelfont=sf,textfont=sf]{subfig}
\usepackage{textcomp}
\usepackage{stfloats}
\usepackage{url}
\usepackage{verbatim}
\usepackage{graphicx}
\usepackage{booktabs}
\usepackage{balance}

\usepackage{booktabs}
\usepackage{enumitem}
\usepackage{subcaption}
\usepackage{titlesec}
\usepackage{booktabs}
\usepackage{multirow}
\usepackage{tabularx}
\usepackage{graphicx} 
\usepackage{MnSymbol}
\usepackage{utfsym}
\usepackage{makecell}
\usepackage{multicol}
\usepackage{bm}
\usepackage{caption}
\usepackage{titlesec}
\usepackage[normalem]{ulem}

\renewcommand{\theparagraph}{\alph{paragraph}}
\titleformat{\paragraph}
  {\normalfont\itshape}   
  {\theparagraph.}         
  {1em}                    
  {}
\titlespacing*{\paragraph}{0pt}{3.25ex plus 1ex minus .2ex}{1em}
\titleformat{\subparagraph}[runin]
  {\normalfont\bfseries}            
  {}                                
  {0pt}                             
  {}
\titlespacing*{\subparagraph}
  {0pt}{1.5ex plus .2ex minus .2ex}{0.5ex plus .1ex}

\biboptions{authoryear}

\begin{document}

\begin{frontmatter}



\title{Survey of End-to-End  Multi-Speaker Automatic Speech Recognition 
for Monaural Audio}


\author{Xinlu~He and Jacob~Whitehill} 

\affiliation{organization={Worcester Polytechnic Institute},
            addressline={100 Institute Road}, 
            city={Worcester},
            postcode={01609}, 
            state={MA},
            country={USA}}

\begin{abstract}
Monaural multi-speaker automatic speech recognition (ASR) remains challenging due to data scarcity and the intrinsic difficulty of recognizing and attributing words to individual speakers, particularly in overlapping speech. Recent advances have driven the shift from cascade systems to end-to-end (E2E) architectures, which reduce error propagation and better exploit the synergy between speech content and speaker identity. Despite rapid progress in E2E multi-speaker ASR, the field lacks a comprehensive review of recent developments.
This survey provides a systematic taxonomy of E2E neural approaches for multi-speaker ASR, highlighting recent advances and comparative analysis. Specifically, we analyze: (1) architectural paradigms (single-input-multiple-output (SIMO) vs.~single-input-single-output (SISO)) for pre-segmented audio, analyzing their distinct characteristics and trade-offs; (2) recent architectural and algorithmic improvements based on these two paradigms, including multi-modal inputs; 
(3) extensions to long-form speech, including segmentation strategy and speaker-consistent hypothesis stitching. Further, we 
(4) evaluate and compare methods across standard benchmarks. We conclude with a discussion of open challenges and future research directions towards building robust and scalable multi-speaker ASR.
\end{abstract}




\end{frontmatter}



\section{Introduction}
Multi-speaker Automatic Speech Recognition (ASR) aims to transcribe speech from audio containing multiple speakers whose speech may overlap. In contrast to single-speaker ASR, which focuses on \textit{what was said}, multi-speaker ASR additionally determines \textit{who says what} \cite{2010Multi-talker__Hershey, settle_2018icsp_end--end_2018, CHIME6}. This task is closely related to the well-known \textit{cocktail party problem} \cite{2018qian_cocktail}, where humans focus on one speaker in a noisy environment filled with competing talkers. Multi-speaker ASR extends this concept by transcribing all speakers in the mixture, and attributing words to individual speakers. This enables practical applications in real-world scenarios such as meetings, group discussions, and phone call recordings, and supports diverse downstream tasks such as meeting summaries, dialogue analytics, and intelligent conversational assistants.

Compared to single-speaker ASR, multi-speaker ASR poses unique challenges, primarily due to overlapping speech and the need for speaker distinction. These require advanced modeling and are further constrained by the scarcity of large, well-annotated datasets. 
Additionally, multi-speaker ASR is inherently multifaceted, involving not only recognition and diarization but also overlap detection, turn-taking detection, and target-speaker ASR. Although these tasks are interrelated and can benefit from joint modeling, effectively leveraging their synergy remains challenging.

While there are multiple comprehensive literature surveys for single-speaker ASR \cite{asr2012,asr2021,asr1}, 
no recent survey has reviewed end-to-end multi-speaker ASR systems. Our paper seeks to fill this gap. 
Before exploring end-to-end solutions, we examine the limitations of traditional cascade architectures, which served as the initial attempts to address the multi-speaker ASR task.

\subsection{The Limits of Cascade Methods} \label{subsec:cascade}

\begin{figure*}[!t]  
    \centering  
    \includegraphics[width=\linewidth]{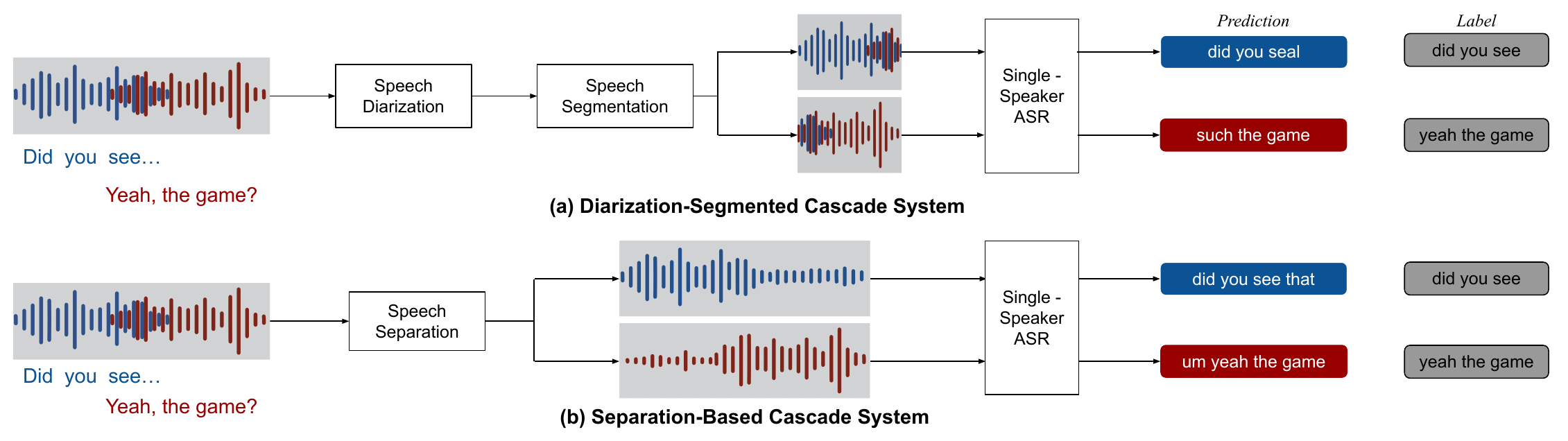}
    \caption{\textit{\textbf{Two Types of Cascade Multi-Speaker ASR Systems.} (a) Diarization-Segmented Cascade System: The mixed audio is segmented by speaker and then processed by a single-speaker ASR model, potentially introducing errors in overlapping speech regions. (b) Separation-Based Cascade System: The mixture is first separated into individual speaker streams using speech separation, followed by single-speaker ASR processing. This method may propagate errors from the separation stage.}}
    \label{fig:modular_methods}
\end{figure*}

Early multi-speaker ASR systems often adopted cascade (modular) methods \cite{2021kanda__comparative_long_term, yu_2022itsp--comparative_2022}, as shown in Fig.~\ref{fig:modular_methods}. One approach is \textbf{diarization-segmented cascade system} (Fig.~\ref{fig:modular_methods}(a)):  (1) Apply speaker diarization to determine \textit{who spoke when} to obtain time-stamped speaker boundaries. (2) Split the audio into segments by detected speaker boundaries. (3) Use a single-speaker ASR model on each segment to obtain individual transcription. This approach leverages well-established single-speaker ASR and works well under minimal overlap. However,  its accuracy degrades with overlapping speech. Moreover, traditional diarization \cite{TradiDiariReview, Sell2014SpeakerDW, Dez2018SpeakerDB, dia12} assumes a single active speaker per frame. Even with later diarization methods that support multi-speaker labeling per frame \cite{Fujita2019EndtoEndNS, Fujita2019Interspeech, ComEmb, ComCluster},  the resulting segments may still contain overlapping speech, posing challenges for single-speaker ASR models.

Another \textbf{separation-based cascade system} (Fig.~\ref{fig:modular_methods}(b)) incorporates a speech separation module. (1) A separation model \textit{enhances} and \textit{denoises} the mixed audio into multiple single-speaker streams, addressing overlapping at the signal level. (2) Each stream is then transcribed using a single-speaker ASR model. While this method can separate overlapping speech in the initial stage, its overall accuracy is dependent on the separation models, which typically optimize signal-level objectives rather than directly targeting ASR performance. Consequently, errors introduced during the speech separation phase can propagate to the subsequent ASR process, compounding the overall error rate of the system \cite{2021kanda__comparative_long_term}. These limitations have motivated end-to-end (E2E) multi-speaker ASR approaches that directly map mixed audio to speaker-attributed transcriptions.

\subsection{Preview of End-to-End Methods}
Unlike cascade methods that explicitly separate speakers before transcription,  E2E systems model \emph{jointly} optimize the complementary problems ``who is speaking'' and ``what is being said''. This can improve accuracy on both tasks. In E2E frameworks, the input is the raw mixed audio, and the output is the transcriptions partitioned by different speakers. 

Initial explorations on E2E multi-speaker ASR \cite{yu_2017--recognizing_2017,seki_2018_acl_a_2018,settle_2018icsp_end--end_2018,chang_2020icsp_end--end_2020} primarily adopted a \textbf{single-input multiple-output} (SIMO) design, where mixed speech is processed through parallel branches to extract speaker-specific representations. These methods typically follow a separation-then-recognition process and are trained end-to-end, often using \textit{permutation invariant training (PIT)} \cite{PIT,PIT2}. A key limitation is that they assume a fixed number of speakers. To address this limitation, \textbf{single-input single-output} (SISO) methods were proposed \cite{kanda_2020itsp--joint_2020, 2021itsp_kanda_e2eTransformer,liang_2023itsp--ba-sot_2023, shi_2024--serialized_2024}, notably using \textit{serialized output training} (SOT) \cite{kanda_2020itsp--serialized_2020}, which generates a unified sequence across speakers. Both SIMO and SISO have since been extended with various architectural and training improvements.

In light of recent advances, this review consolidates recent progress in end-to-end multi-speaker ASR by providing a structured taxonomy of representative models. We analyze core designs and improvements, and compare performance across benchmarks. Three key observations are summarized:
\begin{enumerate}
\item A key distinction in multi-speaker ASR research lies in whether to separate the speech mixture explicitly. Explicit separation generates multiple outputs (SIMO), offering clearer modularity and easier integration with separation and ASR. In contrast, direct mixture processing produces a single output (SISO), preserving contextual information across stages and enabling multi-task learning with mixture-based tasks.
\item There is a growing trend to adapt foundation speech models \cite{Whisper,Wav2Vec,Hubert} to multi-speaker scenarios via lightweight structure modifications and fine-tuning, aiming to mitigate data scarcity. 
\item Model comparison is hindered by limited open-source implementations and inconsistent settings. We provide setting-wise comparisons on standard datasets (AMI, LibriSpeechMix, LibriMix), and analyze models on their design focus.
\end{enumerate}

\subsection{Review Scope and Organization} \label{Setting}
In this paper, we review the recent progress on end-to-end multi-speaker ASR where multiple speakers may talk simultaneously. In particular, we focus on monaural (single-channel) audio recordings and offline (not real-time/streaming) speech analysis.
To structure the comparison of recent models, we identify four key dimensions: 
\textbf{(1) Model architecture:} whether the system follows a SIMO or SISO framework. \textbf{(2) Multi-modal inputs}: whether the model incorporates additional modalities beyond audio; this has become an emerging design choice in recent systems. \textbf{(3) Input granularity:} whether the system is designed and evaluated on pre-segmented clips or long-form continuous audio.
\textbf{(4) Speaker enrollment:} whether the system incorporates speaker enrollment to improve the system.

The rest of the paper is organized as follows:
\begin{itemize}[leftmargin=*,itemsep=0pt,parsep=0pt]
\item Section~\ref{BackgroundTech} reviews background techniques for multi-speaker ASR: end-to-end ASR and speech separation techniques.
\item Section~\ref{sec:Multi-speaker ASR} categorizes and reviews recent advances in SISO and SIMO models for pre-segmented audio.
\item Section~\ref{sec:multimodal} discusses methods that use extra modalities, such as visual information of the speakers' faces, or textual context to make ASR predictions.
\item Section~\ref{long-form} focuses on long-form audio, discussing segmentation techniques and hypothesis stitching.
\item Section~\ref{sec:evaluation} reviews datasets and metrics for multi-speaker ASR, and presents performance comparisons across different experimental settings.
\end{itemize}

\section{Background Techniques} \label{BackgroundTech}
This section reviews two background techniques for multi-speaker ASR. First, we discuss E2E ASR architectures, detailing prevalent model designs and objectives. Second, we outline the speech separation techniques that process mixed audio into isolated streams, which inspire architecture or serve as pre-processing modules for multi-speaker ASR.

\subsection{End-to-End ASR} \label{Background_ASR}
Recent advances in deep learning have enabled E2E network approaches to achieve state-of-the-art performance in ASR. These systems utilize sequence-to-sequence models to directly map speech signals to text outputs. The three primary end-to-end ASR architectures are: (1) Connectionist Temporal Classification (CTC) which relies solely on the previous input; (2) Recurrent Neural Network Transducer (RNN-T), which depends on both previous input and previous output; and (3) Attention-based Encoder-Decoder (AED) architectures, which consider all inputs and previous outputs.

Among these, AED has gained wide appeal through the development of state-of-the-art systems such as Whisper \cite{Whisper}.  In AED, the encoder processes the input acoustic features into a sequence of hidden states, 
and the decoder predicts output tokens 
conditioned on past outputs via an attention mechanism over encoder representations. Early AED implementations relied on RNNs for both the encoder and decoder, but recent models increasingly adopt Transformer~\cite{transformer} and Conformer~\cite{conformer} architectures for their ability to model long-range dependencies with global attention.
Recently, large speech foundation models (e.g., Whisper \cite{Whisper}, Wav2Vec \cite{Wav2Vec}, Hubert \cite{Hubert}) leverage self-supervised learning or multi-task training on massive datasets. After fine-tuning for ASR, these models achieve state-of-the-art performance.

The training objective typically employs a cross-entropy loss to maximize the probability of transcription.
Additionally, a joint CTC loss is often applied to the encoder outputs to enforce monotonic alignment, leveraging its dependence only on previous inputs. This also enhances the encoder's acoustic modeling.
However, due to CTC's strict monotonic constraint, it struggles with overlapping speech, requiring careful adaptation in multi-speaker ASR.

\subsection{Speech Separation Techniques} \label{Background_Separation}
In end-to-end multi-speaker ASR, it sometimes includes a speech separation module and is trained together with the ASR part, inspired by the cascade model. Here we look back at the deep-learning-based speech separation techniques.

The deep-learning-based speech separation can be divided into frequency-domain methods and time-domain methods. The frequency-domain methods process the frequency feature of mixed speech and estimate time-frequency masks or spectral magnitudes for each speaker. The deep clustering framework (DPCL) \cite{DeepClustering} first maps each time-frequency spectral magnitude into a speaker-discriminative embedding, and then the clustering algorithm is used to get the speaker label. Alternatively, a mask output for speech separation can be directly estimated by a deep neural network without embedding. Chimera \cite{Chimera1} combines the two, outputting both speaker embedding and mask. The time domain methods, such as TasNet \cite{TasNet}, directly consume the waveforms using the encoder-separator-decoder framework. The encoder decomposes the mixture into learnable basis functions, the separator estimates speaker-specific weights, and the decoder reconstructs clean waveforms.

In training, label ambiguity arises when multiple outputs must be matched to multiple labels without a predefined order.
Permutation Invariant Training (PIT) \cite{PITseperation2017} addresses this by dynamically aligning model outputs with reference signals. It evaluates all possible permutations and selects the one that minimizes the loss. Consider a mixture of speech from $S$ speakers, and a model produces a set of estimated signals $\{\hat{Y}^s\}_{s=1}^S$. The corresponding reference signals $\{Y^s\}_{s=1}^S$ have no inherent correspondence to the outputs due to the unknown order of speakers. The objective function is defined as:
\begin{equation}
\mathcal{L}_{\text{PIT}} = \min_{\pi \in \mathcal{P}(S)} \sum_{s=1}^{S} \mathcal{L}(\hat{Y}^s, Y^{\pi(s)}),
\end{equation}
where $\mathcal{P}(S)$ represents all permutations of $\{1, \ldots, S\}$, and $\mathcal{L}(\hat{Y}^s, Y^{\pi(s)})$ computes loss between $\hat{Y}^s$ and the permuted reference $Y^{\pi(s)}$.

Despite its effectiveness, PIT faces computational challenges due to the need to evaluate all possible permutations. Additionally, alternative approaches like HEAT~\cite{heat} have been developed to reduce computational cost by using the Hungarian algorithm to efficiently find the optimal permutation.

\section{End-to-End Multispeaker ASR } \label{sec:Multi-speaker ASR}

\begin{figure}[!t]  
    \centering  
    \includegraphics[width=\linewidth]{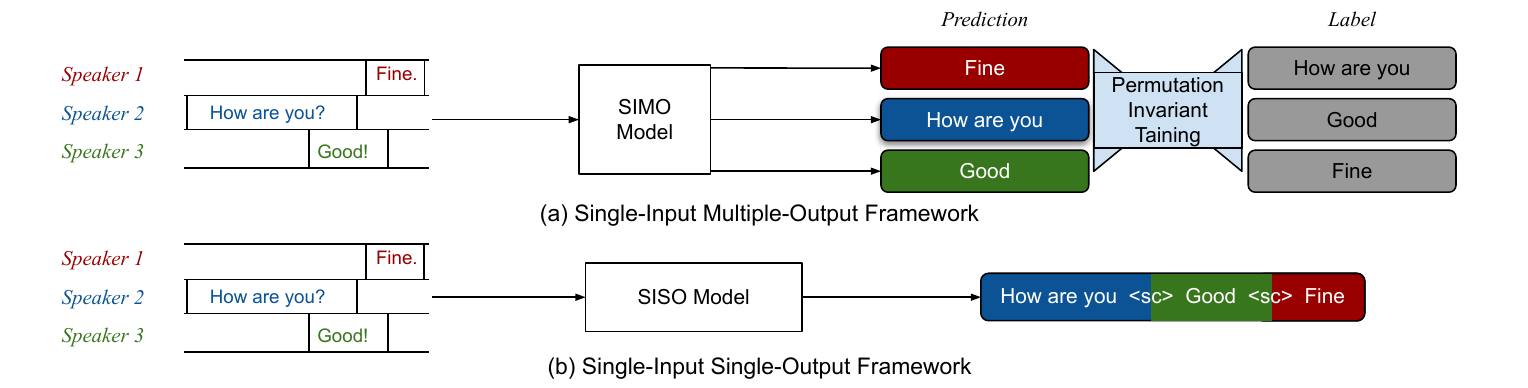}  
    \caption{\textit{Single-input multiple-output (SIMO) and single-input single-output (SISO) processes for multi-speaker ASR.}}  
    \label{fig:SIMO_SISO} 
\end{figure}

\begin{figure}[!t]  
    \centering  
    \includegraphics[width=\linewidth]{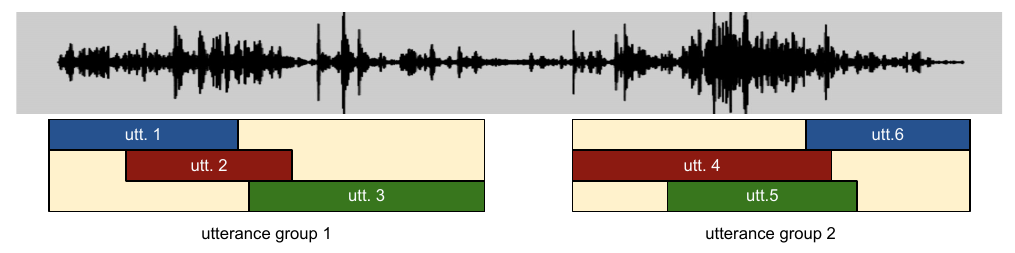}  
    \caption{\textit{Example of utterance groups consisting of overlapping speech segments from multiple speakers. }}  
    \label{fig:utterance_group} 
\end{figure}

In this section, we explore end-to-end multi-speaker ASR systems on two prominent frameworks: single-input single-output 
(SISO) and single-input multi-output (SIMO) (as shown in Fig.~\ref{fig:SIMO_SISO}). While SIMO frameworks generate separate transcriptions for each speaker through parallel branches, SISO frameworks process mixed audio to produce a single transcription output. We review their architectural designs, training methodologies, and key improvements of both frameworks, highlighting their strengths and limitations.
This section focuses on processing predefined audio clips, typically segmented from continuous audio using silence points and heuristic rules (e.g., duration thresholds). 
In single-speaker ASR, such segments are commonly known as \textit{utterances}. For multi-speaker ASR, the concept of \textit{utterance group} \cite{kanda_2021itsp--large-scale_2021} has been introduced as multiple utterances linked through overlapping regions (as shown in Fig.~\ref{fig:utterance_group}).


\subsection{Single-Input Multiple-Output (SIMO)}

SIMO frameworks process mixture audio with exactly $S$ speakers and produce $S$ transcriptions, one per speaker through parallel branches, as illustrated in Fig.~\ref{fig:SIMO_SISO}(a). Here, $S$ is a fixed parameter that denotes the number of speakers. From the standard architecture, we discuss advancements in separation enhancement, speaker scalability, and leveraging pre-trained models.

\subsubsection{Model Architecture} 

The SIMO framework generates a separate transcription for each speaker through distinct output branches. A simple implementation involves a shared network followed by $S$ individual networks, each dedicated to a speaker's transcription. For instance, \cite{yu_2017--recognizing_2017} employs a shared RNN with $S$ linear heads to produce $S$ transcriptions. Alternatively, inspired by separation-based cascade methods, a class of SIMO frameworks (e.g., \cite{seki_2018_acl_a_2018, lin_2022iscslp--separate--recognize_2022}) was proposed to integrate speech separation and ASR in a unified structure.

\begin{figure*}[!t]  
    \centering  
    \includegraphics[trim=.5cm .5cm .5cm .5cm,width=\linewidth]{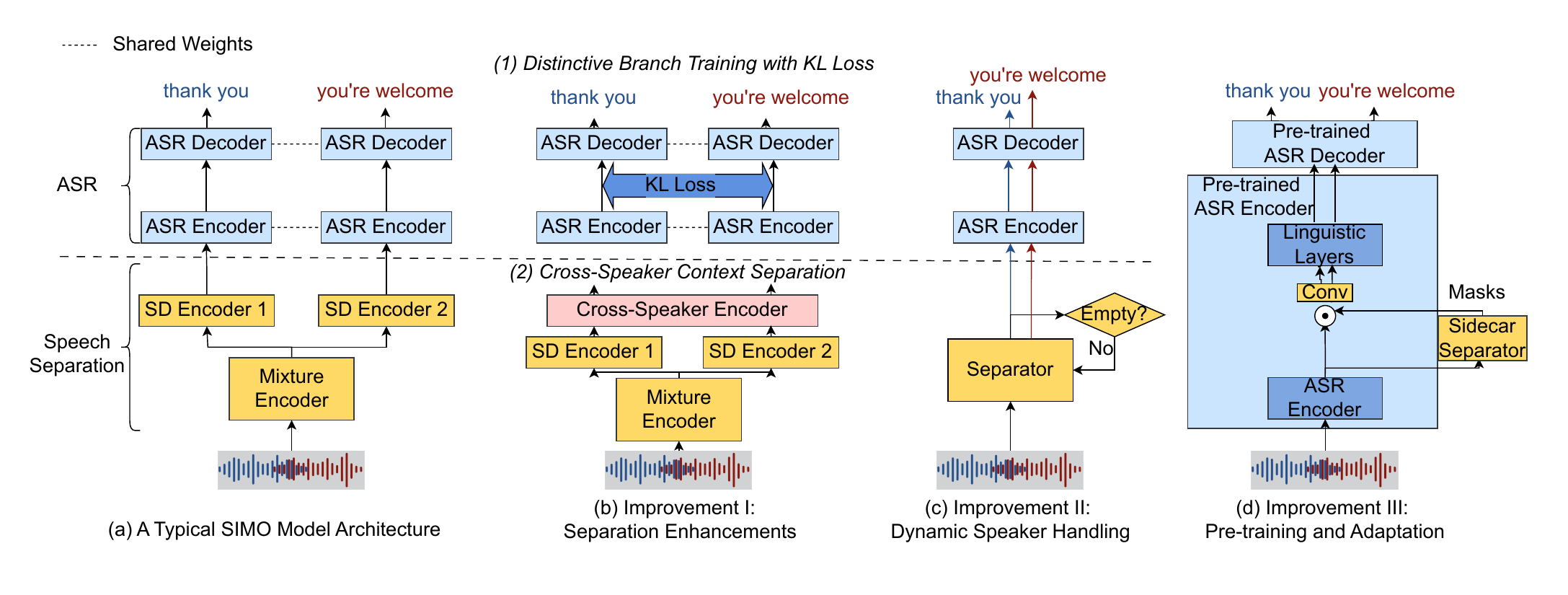}  
    \caption{\textit{\textbf{A typical SIMO architecture (a) and three types of key improvements (b-d).} (b) Enhance separation by (1) Introducing an auxiliary KL Loss to promote distinctive separation between speakers. (2) Incorporating a Cross-Speaker Encoder to provide cross-speaker contextual cues for compensating omission and reducing repetitions between branches; (c) Support dynamic speaker counts using iterative separation. (d) Adapt pre-trained large speech foundation model with a Sidecar separator.} }  
    \label{fig:SIMO_Model} 
\end{figure*}

Fig.~\ref{fig:SIMO_Model}a illustrates a typical SIMO model architecture \cite{seki_2018_acl_a_2018,chang_2020icsp_end--end_2020,zhang_2019itsp--knowledge_2019,zhang_2020tasl--improving_2020} that integrates the speaker separation process and ASR into a single framework, enabling end-to-end training from scratch. The architecture adopts a stacked design comprising shared and unshared modules across branches. The process begins with a mixture encoder, $\text{Encoder}_{\text{Mix}}$, which extracts an intermediate feature sequence $\mathbf{Z}$ from the mixed audio input $\mathbf{X}$, serving as the input for subsequent separation and ASR. This feature sequence is then fed into $S$ parallel branches, each dedicated to one of the $S$ speakers. Within branch $s$, a speaker differentiating encoder, $\text{Encoder}_{\text{SD}^s}$, disentangles the speech content $\mathbf{Z^s}$ of the corresponding speaker from the mixture feature $\mathbf{Z}$. Finally, a shared ASR model, such as an attention-based encoder-decoder (AED) or transformer, generates the transcription hypothesis $\mathbf{H^s}$ for speaker $s$. The process can be formally described as:
\begin{align}
\mathbf{Z}&= \text{Encoder}_{Mix}(\mathbf{X}), \\
\mathbf{Z}^s &= \text{Encoder}_{SD^s}(\mathbf{Z}), \quad s = 1, \dots, S\\
\mathbf{H}^s &= \text{ASR}(\mathbf{Z}^s), \quad s = 1, \dots, S
\end{align}

The training objective of each branch follows that of single-speaker ASR: cross-entropy loss $\mathcal{L}_{\text{CE}}$ to maximize the transcription accuracy, and an optional CTC loss $\mathcal{L}_{\text{CTC}}$ for monotonic alignment. Similar to speech separation, SIMO models with multiple output branches introduce label ambiguity, where the correspondence between hypotheses and references is unclear. Permutation Invariant Training (PIT) is also utilized to align hypotheses $\mathbf{H}^s = (h_1^s, \ldots, h_{N_s}^s)$ with reference transcriptions $\mathbf{R}^{s} = (r_1^{s}, \ldots, r_{N_{ss}}^{s})$ through permutations \cite{yu_2017--recognizing_2017}. 
To reduce the permutation computational cost, label matching is often determined solely based on the CTC loss, while both CTC and cross-entropy losses are used (\cite{seki_2018_acl_a_2018,chang_2019icspend--end_2019,zhang_2019itsp--knowledge_2019,zhang_2020tasl--improving_2020}).

SIMO provides a natural framework for integrating traditional separation and ASR methods into an end-to-end system. This design facilitates the incorporation of state-of-the-art separation techniques to enhance performance. However, the fixed number of branches limits the model's ability to handle a variable number of speakers, which remains a significant constraint in real-world scenarios. Also, limited separation performance still causes redundancy and omissions in the final transcription output.

\subsubsection{SIMO Improvements}
\begin{figure}[!t]  
    \centering  
    \includegraphics[width=\linewidth]{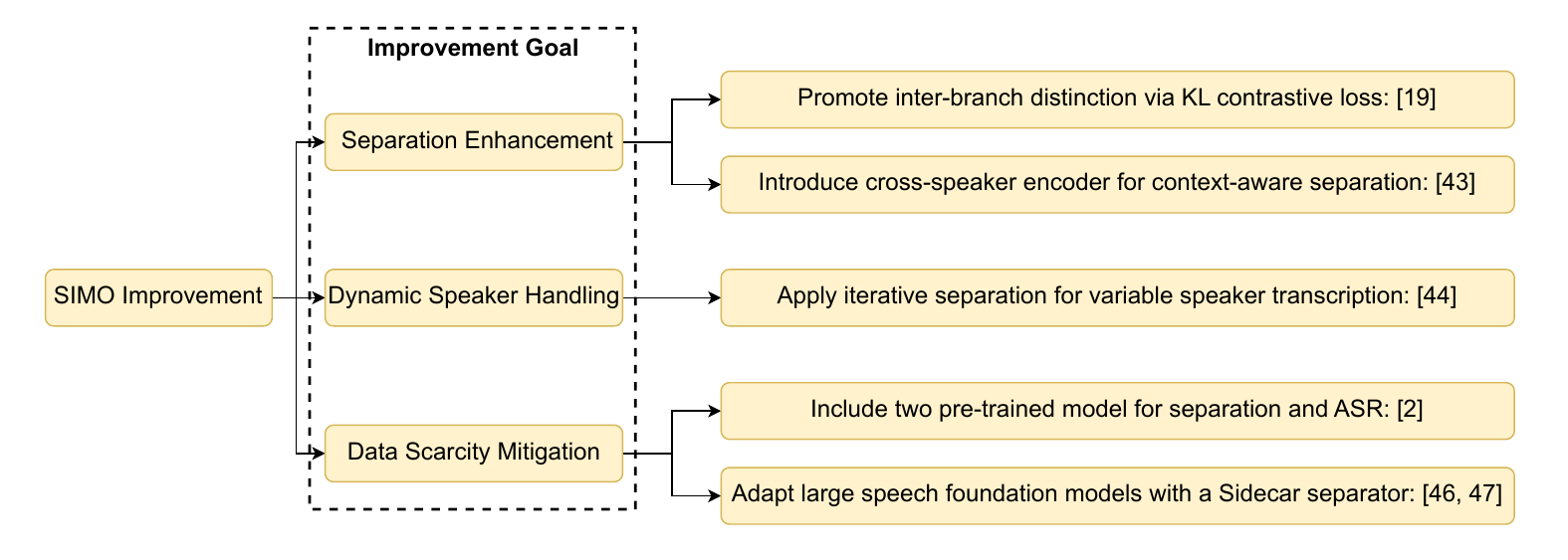}  
    \caption{\textit{\textbf{SIMO Improvements by Goal}. Targeting better separation performance, variable speaker handling, and data scarcity mitigation by module refinement, iterative processing, and pre-trained model integration. }}  
    \label{fig:SIMO_Improvements} 
\end{figure}

To address SIMO’s limitations, recent research has focused on three key goals: (1) enhancing separation performance, (2) enabling a flexible number of speakers, and (3) mitigating data scarcity. The following sections detail these categories. Fig.~\ref{fig:SIMO_Improvements} summarizes the corresponding goals and methods.

\paragraph{Separation Enhancement} 

While end-to-end models jointly train separation and recognition modules, independent branches can propagate early separation errors, resulting in repeated or omitted transcriptions. Recent work addresses this by increasing inter-branch distinctiveness and leveraging context to refine separation.

\subparagraph{Distinctive Branch Training: }
To encourage distinct transcriptions across output branches, auxiliary loss functions can be applied between ASR hidden state vectors. In AED-based models, Seki et al. \cite{seki_2018_acl_a_2018} proposed a contrastive loss that maximizes Kullback-Leibler (KL) divergence between the ASR encoder states from different branches, as shown in Fig.~\ref{fig:SIMO_Model}(b.1). 

This loss penalizes similarity between streams, reducing the likelihood of redundant transcriptions. 

\subparagraph{Context-Aware Separation:}
More recently, the separation has been further improved by introducing cross-speaker context to enhance the separation quality. Traditional SIMO systems process each speaker independently in parallel branches, restricting the model’s capacity to capture cross-speaker dependencies and perform mutual correction. To address this, Kang et al. \cite{kang_2024icsp--cross-speaker_2024} proposes a \textit{Cross-Speaker Encoder} ($\text{Encoder}_{\text{CSE}}$), positioned between the $\text{Encoder}_{SD}$ and ASR model to enable information sharing across branches (Fig.~\ref{fig:SIMO_Model}(b.2)). It consumes the concatenation of mixture encoding from $\text{Encoder}_\text{{Mix}}$ and different speaker's individual encodings from $\text{Encoder}_\text{{SD}}$, and employs a conformer block to enable information sharing across branches. The conformer output is then partitioned into branch-specific representations and passed to the respective ASR encoders. This context-aware mechanism leverages the local context of the mixed speech to refine the single-speaker features, thereby improving separation accuracy.

\paragraph{Dynamic Speaker Handling} 
A key limitation of traditional SIMO frameworks is their reliance on a fixed number of speaker streams, which constrains their applicability to real-world scenarios. Recent work addresses this challenge by introducing iterative and adaptive systems to determine dynamically the number of speakers. As illustrated in Fig.~\ref{fig:SIMO_Improvements}(c), Neumann et al. \cite{neumann_2020itsp--multi-talker_2020} propose an iterative approach where the system separates one speaker’s audio at a time with a Dual-Path RNN TasNet \cite{Dual-Path-RNN} separator. The residual mixture is then passed to subsequent iterations until no speech remains. Each separated feature is then fed into a shared ASR model, which can be jointly fine-tuned with the separator. This design supports end-to-end training while supporting varying numbers of speakers.
\paragraph{Pretraining and Adaptation}
Leveraging pre-trained models mitigates the data scarcity challenge in mult-speaker ASR, as shown in Fig.~\ref{fig:SIMO_Model}(d).
In the SIMO paradigm, a common approach is to adopt a dual-module framework consisting of a pre-trained speech separation module and a pre-trained ASR module, with joint fine-tuning to align their objectives. For example, Settle et al. \cite{settle_2018icsp_end--end_2018} includes Chimera++, a pre-trained separation module, to generate speaker masks over mixture features, producing \(S\) separate streams. These streams are then fed into a shared ASR model for transcription, and the two modules are subsequently fine-tuned together to enable end-to-end optimization.

More recent advances introduce large speech foundation models like Whisper \cite{Whisper} and Wav2Vec \cite{Wav2Vec} into the SIMO pipeline.
Since these models are not inherently designed for multi-speaker inputs, a separation module must be inserted to enable SIMO-style processing.
To this end, Meng et al. \cite{meng_2023ICSP_sidecar, meng_2024--empowering_2024} propose a lightweight convolutional network Sidecar separation module inserted between layers of a frozen Wav2Vec2.0 or Whisper model. This design splits the mixed audio into multiple streams, each processed independently by subsequent Wav2Vec layers to produce distinct transcriptions. Instead of building an external separation module, it operates directly on the internal feature representation of the pre-trained model, enabling an efficient and seamless adaptation of pre-trained single-speaker ASR models to multi-speaker scenarios.

\subsection{Single-Input Single-Output (SISO)}
SISO is another multi-speaker ASR framework that only generates a single output sequence containing all speakers' transcriptions. Unlike SIMO, which produces separate outputs per speaker, SISO relies on serialized output training (SOT) \cite{kanda_2020itsp--serialized_2020} to serialize transcriptions into a unified stream. The transcriptions of all speakers are concatenated into one output. 

SISO offers several advantages. First, it naturally handles varying numbers of speakers, making it well-suited to real-world scenarios with unknown speaker counts. Second, by generating a single output sequence, SISO captures inter-speaker dependencies, improving both coherence and overall transcription accuracy. Third, it reduces computational cost by enforcing a fixed output order, thereby simplifying training.

This section first introduces
two forms of SOT output: speaker-ordered sentence-based SOT and temporal-ordered token-based SOT. 
We then discuss improvements to the SISO architecture.

\subsubsection{Serialized Output}
\begin{figure}[!t]  
    \centering  
    \includegraphics[width=\linewidth]{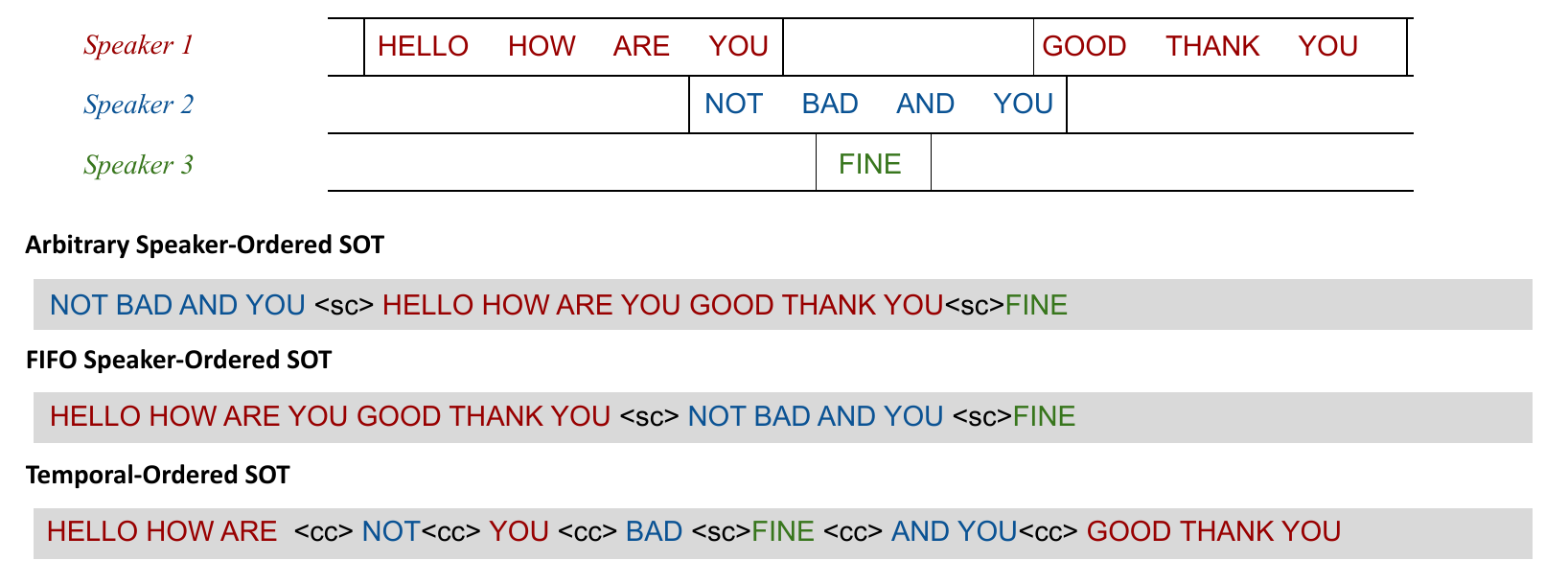}  
    \caption{\textit{An example of different SOT texts, containing speaker-change symbol \texttt{<sc>} and \texttt{<cc>}.}}  
    \label{fig:SIMO_SOT_Example} 
\end{figure}

SOT can be implemented in two primary forms: sentence-based SOT and token-based SOT (Fig.~\ref{fig:SIMO_SOT_Example}). In sentence-based SOT, all sentences of each speaker are concatenated into a sequence. Special token $\texttt{<sc>}$ is inserted between the transcriptions to indicate changes in the speaker. Consecutive sentences from the same speaker are simply concatenated in the transcript, even if those sentences may be partially or even completely interrupted by another speaker's sentence. The order of speakers can be arbitrary, with PIT used in loss.
Alternatively, sentence-based SOT can follow a first-in, first-out (FIFO) order, where transcriptions are arranged by each speaker's start time, from the earliest to the latest. 

In contrast, token-based SOT serializes transcriptions based on token timestamps. The channel change symbol \texttt{<cc>} is utilized in token-based SOT. This approach provides finer granularity in capturing overlapping speech by preserving the timing order, while remaining compatible with CTC loss, which is well-suited for mostly time-monotonic sequences \cite{li_2024icsp--improving_2024} \cite{fan_2024icsp-sa-sot_2024}.
However, when the same speaker's sentence is interleaved with others, a more advanced decoder is needed to effectively model long-range context and maintain coherence within that speaker's speech.

\subsubsection{SISO Model and Improvements}
\begin{figure*}[!t]  
    \centering  
    \includegraphics[trim=.5cm .5cm .5cm .5cm,width=\textwidth]{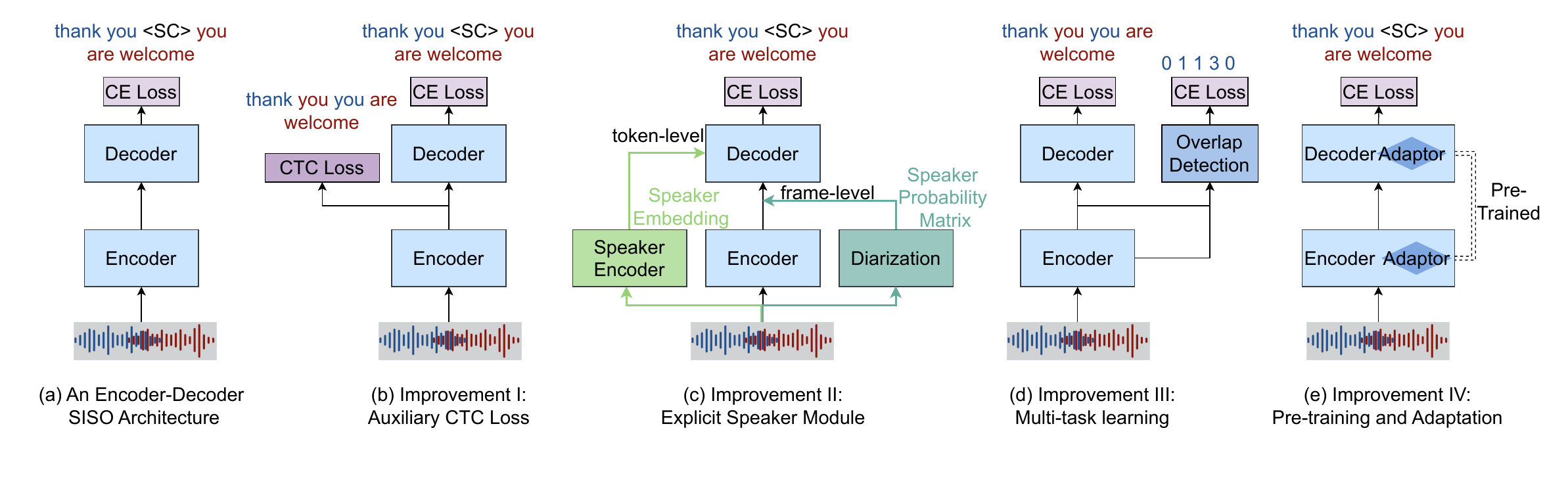}  
    \caption{\textit{\textbf{An encoder-decoder SISO architecture (a) and four representative improvements (b-e)}. (b) Adding an auxiliary CTC loss to enhance acoustic modeling and speaker-awareness capability; (c) Incorporating explicit speaker modules to inject speaker information at the frame or token level; (d) Applying multi-task learning with auxiliary tasks such as overlap detection; (e) Integrating pre-trained models and fine-tuning them with mechanisms such as adapters.}}  
    \label{fig:SISO_impFig} 
\end{figure*}

Since SISO models produce a single output stream, they typically build on single-speaker architectures without requiring explicit speaker separation.
Fig.~\ref{fig:SISO_impFig}(a) shows an encoder-decoder model that has been successfully adapted to SISO-based multi-speaker ASR in \cite{kanda_2020itsp--serialized_2020}. In this setting, the ASR encoder generates hidden representations $\mathbf{Z}^\text{asr}$, which are then passed to ASR decoder to obtain serialized transcriptions.

\begin{figure*}[!t]  
    \centering  
    \includegraphics[width=\textwidth]{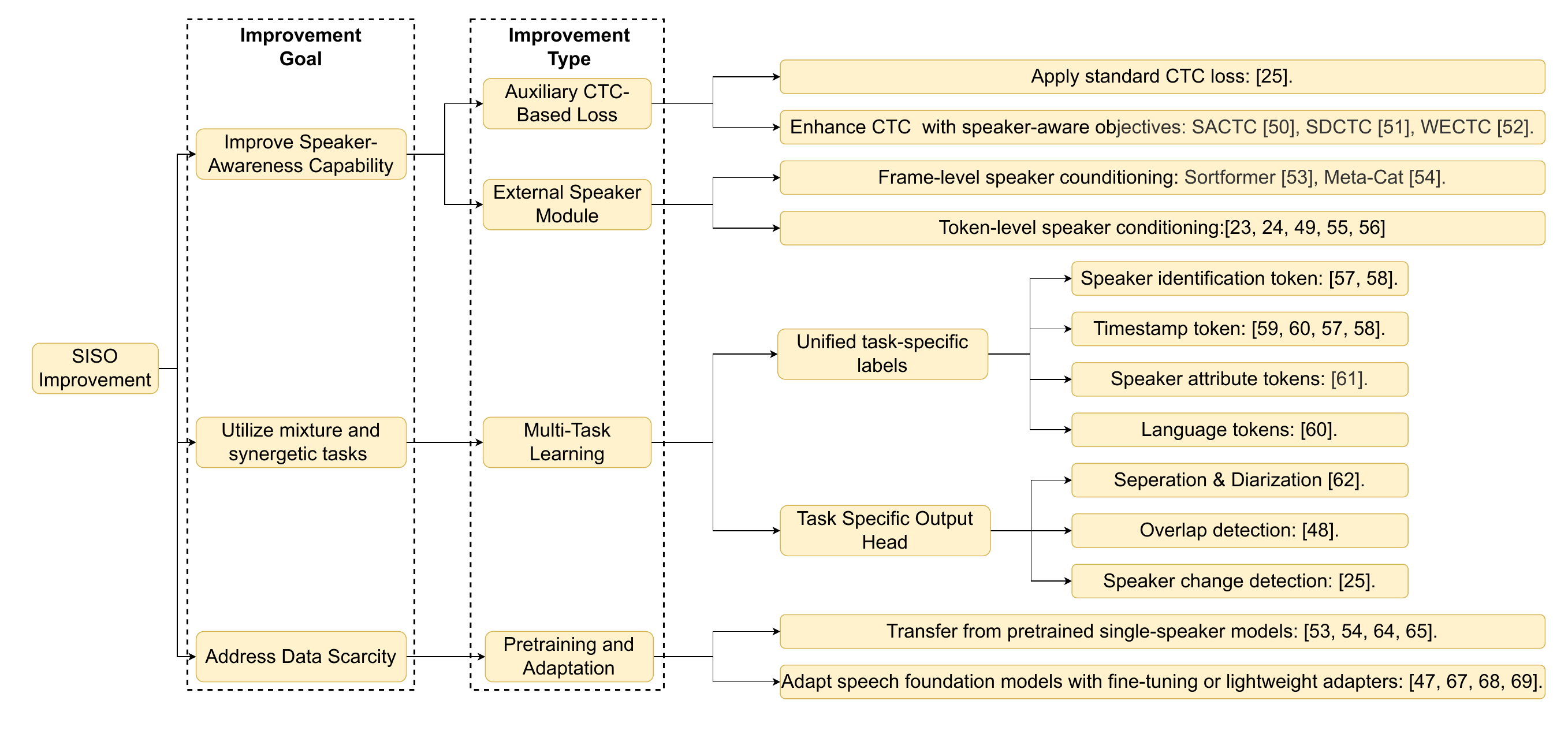}  
    \caption{\textit{\textbf{SISO Improvements by Goal and Type}. Aiming to enhance speaker-aware capabilities, utilize shared representation and alleviate data scarcity via auxiliary losses, external speaker modules, multi-task learning, and pre-trained models adaptation.}}  
    \label{fig:SISO_Improvements} 
\end{figure*}

Although using a single processing path enables cross-speaker context modeling, the lack of explicit separation mechanisms in SISO systems makes accurate transcription of overlapping speech challenging.
To address this and compensate for limited multi-speaker training data, researchers have proposed four main strategies: (1) auxiliary CTC-related losses, (2) integration of external speaker modules, (3) multi-task learning, and (4) pre-training and adaptation.
Figure \ref{fig:SISO_Improvements} shows the improvement type.

\setlist[itemize]{
    noitemsep,          
    topsep=0pt,         
    partopsep=0pt,     
    leftmargin=*,       
    after=\vspace{-\baselineskip} 
}

\paragraph{Auxiliary CTC-Related Loss}

As described in Section \ref{Background_ASR}, CTC loss commonly serves as an auxiliary objective to cross-entropy loss in AED to enhance the encoder's acoustic modeling. 
In the SISO framework, it can be applied similarly, through a parallel branch connected to the encoder output, independent of the attention-based decoder, as illustrated in Fig. \ref{fig:SISO_impFig}.
In addition, modified CTC variants can incorporate speaker information to improve speaker differentiation in multi-speaker transcription, as described below. 

\subparagraph{CTC in SISO:}
To leverage CTC for improving acoustic alignment in SISO systems, the CTC loss objective must maintain temporal consistency with the original mixed acoustic features. Typically, token-based SOT serves as the CTC target in this setting. If the model’s final output follows sentence-based SOT, the remaining components (not supervised by CTC) need to handle the utterance-level reordering.
For instance, Liang et al. \cite{liang_2023itsp--ba-sot_2023} proposed a two-stage CTC approach: the first CTC loss is applied after the initial encoder layers, optimized for token-based SOT to enhance acoustic modeling, while the second CTC loss operates on the full encoder output with sentence-based SOT supervision.

\subparagraph{Speaker-Enhanced CTC: }
The second approach modifies CTC to explicitly integrate speaker information. A noticeable example is Speaker-Aware CTC (SACTC) \cite{kang_2024--disentangling_2024} proposed by Kang et al., which formulates a Bayes-risk-based CTC that constrains the encoder to distinguish speaker-specific features at specific time frames, explicitly modeling speaker separation. More recently, Speaker-Disguishable CTC (SD-CTC)~\cite{sdctc} extends CTC by jointly assigning a token and its corresponding speaker label to each frame. Another advancement comes from Zheng et al. \cite{zheng_2024--unsupervised_2024} with  Weakening and Enhancing CTC (WECTC) loss, which enhances speaker change token (\texttt{<sc>}) prediction by adjusting pseudo-label posteriors.

Despite its widespread use in SISO frameworks and extensions with speaker-enhanced modules, CTC remains limited in overlapping speech scenarios, as shown in the results of \cite{kang_2024--disentangling_2024}. It assumes conditional independence between output tokens and enforces a one-token-per-frame constraint, producing a single, linear output sequence. This makes it fundamentally incompatible with simultaneous speech from multiple speakers, often resulting in entangled or incomplete transcriptions. Future research may investigate approaches that relax this constraint while preserving CTC's alignment ability.

\paragraph{External Speaker Module}
One direction for improving the SISO framework is the integration of explicit speaker information into the model architecture. By incorporating speaker-specific cues, the model can enhance its ability to differentiate between speakers and better handle overlapping speech. Current approaches can be broadly categorized into two methodologies: (1) Frame-level Speaker Conditioning (FSC) and (2) Token-level Speaker Conditioning (TSC). 

\subparagraph{Frame-level Speaker Conditioning (FSC)}
incorporates speaker information by aligning speaker and speech timestamps at the frame level.
Modular FSC \cite{yu_2022itsp--comparative_2022} directly aligns diarization results with SOT transcriptions in a post-processing stage, which is not an end-to-end approach and may lead to error propagation.
Later end-to-end FSC methods (right side of Fig.~\ref{fig:SISO_impFig}(c)) integrate speaker information before the decoder, aligning the diarization output with ASR encoding representation at the frame level.
Specifically, the diarization module generates an $S$-speaker assignment probability matrix for $T$ frames, denoted as $\mathbf{P} \in \mathbb{R}^{S\times T}$.
This matrix is used to incorporate speaker information into the ASR encoder representations $\mathbf{Z} \in \mathbb{R}^{M \times T}$ before passing the fused representation to the decoder through different integration strategies. 
One approach \cite{park_2024--sortformer_2024} incorporates a sinusoidal matrix, similar to the sinusoidal positional encoding used in Transformers \cite{transformer}. Each speaker is associated with a unique sinusoidal pattern, which is then weighted by the probability matrix $\mathbf{P}$ and added to the original encoder representation $\mathbf{Z}$. This mechanism can differentiate speakers in a structured, deterministic way, and can be disabled to fall back to a standard ASR pipeline.
Alternatively, Wang et al. proposed Meta-Cat \cite{wang_2024--meta-cat_2024}, which first computes a speaker-specific representation $\mathbf{Z}_s$ by element-wise multiplying the encoder output $\mathbf{Z}$ with the frame-level probability vector $P_s$ for each speaker $s$.
These representations are then concatenated across all speakers to form a supervector, which is passed to the decoder. This method expands the ASR embedding multiple times according to the number of speakers.
Both approaches ensure the entire model remains differentiable, enabling joint training of the whole system.

\subparagraph{Token-level Speaker Conditioning (TSC)}
introduces speaker embeddings as auxiliary input to the decoder during token generation. Unlike FSC, which aligns frame-level probabilities, TSC typically uses a speaker encoder to extract speaker embeddings for speaker-aware decoding (left side of Fig.~\ref{fig:SISO_impFig}(c)).  
A typical TSC speaker-attributed ASR system (\cite{kanda_2020itsp--joint_2020, kanda_2021slt_investigation_2021, 2021itsp_kanda_e2eTransformer}) consists of four modules: ASR encoder, speaker encoder, speaker decoder, and ASR decoder. The input mixture is processed in parallel by the ASR and speaker encoders to generate the speech representation $\mathbf{Z}^\text{asr}$ and speaker embeddings $\mathbf{Z}^\text{spk}$, respectively. These are fed into the decoder, along with the previous token sequence, to produce the context-aware speaker representation. The ASR decoder then takes this speaker representation as extra input with ASR encoder states and previous output to predict the next token. 

Here, the speaker representation is related to a pre-defined speaker inventory $\mathcal{D} = \{\mathbf{d}_1, \dots, \mathbf{d}_K\}$, where $\mathbf{d}_k$ represents a speaker profile in the inventory. The output of the speaker decoder is used as a query to compute attention over this speaker inventory $\mathcal{D}$, and generates a weighted speaker profile $\bar{\mathbf{d}}_n$ as final speaker representation given to the speaker decoder.
Later approaches proposed by Shi et al. (\cite{shi23d_interspeech}) considered contextual information for speaker representation generation. A separate contextual text encoder is deployed before the speaker decoder to aggregate the semantic information of the whole output utterance. In addition, when calculating $\bar{\mathbf{d}}_n$, an extra context-dependent scorer is employed to model the local speaker discriminability by contrasting with speakers in the context. Fan et al. \cite{fan_2024icsp-sa-sot_2024} also enhance the speaker contextural relationship by replacing the weighted profile $\bar{\mathbf{d}}_n$ as a similarity matrix and passing this matrix to the ASR decoder. In this way, the system can incorporate the speaker information without the speaker inventory $\mathcal{D}$.

\paragraph{Multi-Task Learning} \label{subsubsubsec:multi-task}
Multi-task learning enables the joint optimization of synergistic tasks via a fully or partially shared network, which can in turn improve the performance of individual tasks (e.g., multi-speaker ASR and overlapping detection). Compared to its application in SIMO models, multi-task learning in SISO can directly leverage the shared representation and jointly train the ASR model with mixture-related tasks such as diarization and overlap detection by sharing network components. 

\subparagraph{Unified Labeling: } Task-specific labels can be inserted into the serialized output as special tokens. This unified labeling enables joint modeling of multiple tasks, such as speaker identification and timestamp predictions, without modifying the original auto-regressive multi-speaker ASR architecture. First, specific speaker tokens (e.g., \texttt{<spk0>} \texttt{<spk1>}) can replace general speaker changing tokens (\texttt{<sc>} and \texttt{<cc>}) to differentiate speakers, as in \cite{cornell_2024icsp--one_2023} \cite{makishima_2024itsp--somsred_2024}. This reduces the speaker ambiguity in SOT, especially token-level SOT. Second, quantized timestamp tokens (e.g., 20ms resolution) can be inserted as special tokens to indicate the start and end time of an utterance \cite{makishima_2023itsp--joint_2023,li_2023z--adapting_2023,cornell_2024icsp--one_2023,makishima_2024itsp--somsred_2024}. 
Timestamp tokens provide extra information about speaker turn-taking, temporal order, and overlapping speech. In some systems, additional timestamp tokens are inserted based on heuristic rules, such as when the gap between consecutive tokens exceeds two seconds, to indicate pauses or silence \cite{li_2023z--adapting_2023}.
Additionally, Masumura et al. \cite{masumura_2021--unified_2021} and Li et al. \cite{li_2023z--adapting_2023} independently explored the use of speaker attribute tokens (e.g., gender, age) and language tokens, respectively, to enrich the output with contextual information. 
All tasks under this unified labeling scheme are handled by a single model with a shared output head.

\subparagraph{Task Specific Output Head: }
Separate labels and output heads for each auxiliary task can serve as another implementation of multi-task learning in SISO. In this case, tasks share parts of the network, typically lower-layer acoustic or speaker-related representations, while maintaining task-specific output branches in the later stages. For example, Shakeel et al.~\cite{shakeel2025unifying} introduce a unified multi-speaker encoder (UME) that jointly addresses the speech separation, speech diarization, and multi-speaker ASR tasks within a single architecture. Their approach uses the Open Whisper-style Speech Model (OWSMv3.1)~\cite{OWSM} to extract representations from multiple layers, and fuses them using learnable weighting parameters. The resulting shared representation is then fed into task-specific heads. For the separation module, the model also directly incorporates the original speech mixture as an additional input. Overall, UME demonstrates substantial gains over strong single-task SOTA baselines across three tasks.
Li et al. \cite{li_2024icsp--improving_2024} jointly optimize the overlapping prediction task and the ASR task to enhance multi-speaker ASR (Fig.~\ref{fig:SISO_impFig}(d)). Specifically, the overlap-aware task predicts two binary states for each token: whether it overlaps with other tokens and whether it is a boundary token. Both the ASR and overlapping prediction models adopt an encoder-decoder structure, sharing lower layers of the encoder to leverage acoustic features while diverging to predict overlap-aware labels. The total loss combines the ASR loss and the overlap-aware loss, enabling joint training to improve performance in overlapping speech scenarios.

Speaker change detection has also been integrated as an auxiliary task in multi-task learning frameworks. For example, Liang et al. \cite{liang_2023itsp--ba-sot_2023} proposed Boundary-Aware SOT (BA-SOT). Specifically, the model adds a speaker change detection (SCD) module as the output head for the speaker change task after several decoder blocks. A binary speaker change label is used for the auxiliary task, while the ASR task adopts FIFO-style SOT labels without speaker change tokens. The total loss consists of ASR loss, SCD loss, and a newly introduced boundary constraint loss.
Studies \cite{liang_2023itsp--ba-sot_2023, li_2024icsp--improving_2024} have shown that  joint modeling of ASR with auxiliary tasks such as speaker change detection and overlap detection can lead to improved recognition accuracy.

\paragraph{Pretraining and Adaptation}
Similar to SIMO, leveraging pre-trained modules in SISO architectures helps mitigate the data scarcity in multi-speaker ASR. 
Because SISO models share the same model structure and single-output format with single-speaker ASR, SISO models can be directly initialized from pre-trained single-speaker models without architectural modification \cite{denisov_2019--end--end_2019, rose2023cascadedencodersfinetuningasr}. This facilitates transfer learning from large-scale simulated mixtures, which are artificially created multi-speaker audio samples (see section \ref{sec:evaluation}). These simulated datasets can be used for pre-training and subsequently fine-tuned on real-world corpora to achieve competitive performance \cite{kanda_2021itsp--large-scale_2021}. Moreover, speaker-specific modules in SISO models -- such as speaker encoder and diarization model -- can also be initialized from pre-trained modules \cite{park_2024--sortformer_2024,wang_2024--meta-cat_2024}.

This structural compatibility also holds for speech foundation models: single-speaker models, such as Whisper and WavLM-based ASR, can be directly fine-tuned under the SISO framework for multi-speaker ASR, unlike SIMO models that require additional separation modules. These models can either be fully fine-tuned, or adapted using lightweight modules like LoRA~\cite{hu2022lora}, enabling resource-efficient training while keeping most pre-trained parameters frozen \cite{shi2024advancingmultitalkerasrperformance,wang_2024z--resource-efficient_2024,meng_2024--empowering_2024, polok_dicow_2026}. 
In addition, Shi et al. \cite{shi2024advancingmultitalkerasrperformance} explored maintaining the multilingual property of foundation models by leveraging adapters.

\subsection{SIMO vs.~SISO: Summary}\label{subsec:SIMO/SISO_Summary}
By performing speaker separation and speech recognition separately, SIMO provides more modularity than SISO, which performs both tasks in an integrated way. However, SIMO is less flexible than SISO in three key aspects: (1) SISO accommodates variable speaker counts, whereas SIMO requires knowing the number of speakers \emph{a priori}; (2) SIMO's branch-specific separation lacks cross-speaker validation for redundancy and complementarity and further underutilizes cross-speaker information in ASR. Very  few (e.g., \cite{kang_2024icsp--cross-speaker_2024}) SIMO approaches attempt to address this shortcoming.
(3) SISO's retention of mixed features enables  multi-task learning with mixture-based tasks, such as overlapping detection. Additionally, SISO's fixed speaker order in transcription (e.g, FIFO) reduces computational overhead from label matching in SIMO training.

These structural differences lead to distinct improvement priorities. SIMO's overall recognition performance is fundamentally limited by its separation module: poor separation introduces residual noise or truncated phonemes, degrading ASR accuracy. Thus, SIMO requires enhanced separation for optimal performance. In contrast, SISO lacks explicit separation, and thus harnessing speaker information to disambiguate overlapping speech is a focus for SISO improvement. Some advances are largely incremental, such as variations of the CTC loss or other light-weight architectural refinements. In contrast, other developments, such as the incorporation of large foundation models, represent more transformative progress in multi-speaker ASR.

Both SIMO and SISO address data scarcity by utilizing pre-trained models, particularly recent large foundation models. For SIMO, this requires adding a separation process to the foundation model.
SISO directly fine-tune the foundation ASR model on multi-speaker data.
Both frameworks can achieve strong performance by tuning only 8-10\% of all parameters.

\subsection{Future Directions}\label{subsec:Future}
One promising direction is \textbf{SIMO-SISO hybridization}, which aims to combine SIMO’s separation precision with SISO’s comprehensive modeling. In such framework, the model can first separate the input audio into multiple speaker-specific branches, whose representations are then concatenated and passed to additional network layers for integrated decoding. Recent work has explored this hybridization at different representation levels. For example, Cross-speaker encoder \cite{kang_2024icsp--cross-speaker_2024} combined  acoustic features from different branches before ASR, to enhance separation by providing cross-speaker context.
Also, Huang et al. \cite{huang_2023icsp--adapting_2023} adapt WavLM to extract target-speaker transcriptions from mixture audio using speaker embeddings. The resulting utterances are concatenated and passed through a joint speaker module to generate the final serialized transcript. Future studies may further explore multi-level integration with advanced model architecture to enhance SIMO’s contextual modeling and improve SISO’s ability to handle overlapping speech.

Another direction is the \textbf{adapted foundation model enhancement}. Foundation-model adaptation represents a transformative direction with the potential to reshape multi-speaker ASR. There are lots of questions that remain underexplored. For instance, can a foundation-adapted SIMO model also leverage cross-speaker context? How can foundation-adapted SISO models benefit from multi-task learning? Beyond architecture-specific adaptations, future work may further investigate the integration of multi-speaker ASR into multi-modal LLM-based unified frameworks. This can extend current instruction-based approaches \cite{shi_2024--advancing_2024} to more practical, real-world applications.

\section{Multi-Modal Multispeaker ASR}
\label{sec:multimodal}
The previous section categorized end-to-end multi-speaker ASR systems primarily based on their output format. However, from the input perspective, a growing body of work explores feeding multi-speaker ASR models with multi-modal inputs. When the acoustic mixture is excessively noisy, is highly overlapped, or contains rare or long-tail words that challenge audio-only systems, additional modalities can provide complementary cues to better resolve both speaker identity and linguistic content.
\\
\indent Most existing multi-modal approaches follow the   SISO architectures, as this design retains the complete acoustic mixture throughout the encoder and decoder, thus allowing high-level visual or textual context to be fused directly into the joint modeling of speakers and speech. In contrast, SIMO pipelines must decide where to incorporate cross-modal signals: either before separation or after separation.
\\
\indent The auxiliary context usually falls into two categories: (1) visual information, such as video and images; (2) textual information that leverages the text-understanding capabilities of multi-modal LLMs.
\subsection{Audio-Visual Multi-speaker ASR}
Visual information in audio-visual speech recognition typically comes from face tracks and mouth movements. Prior single-speaker studies \cite{avasr1,avasr2,avasr3_Seo_2023_CVPR} have consistently shown that visual cues can improve ASR robustness, especially under adverse acoustic conditions, because the visual modality is unaffected by background noise. In most systems, visual features are incorporated by concatenating them with audio features, which is then given to audio-visual ASR model.
\\
\indent In multi-speaker scenarios, visual cues also help improve transcription accuracy. Early multi-person audio-visual ASR studies \cite{2020avsr, braga2022closerlookaudiovisualmultiperson} explored how to incorporate face-track features into a standard ASR model without explicitly predicting speaker labels. Importantly, these approaches do not require any training labels regarding which face corresponds to which voice; rather, this correspondence is inferred dynamically by the model. Let $\mathbf{Z}^{a}$ denote the audio encoder representation and $\mathbf{Z}^{v_s}$ the visual representation for speaker $s \in \{1,\ldots,S\}$. A cross-attention module takes $\mathbf{Z}^{a}$ as the query and $\{\mathbf{Z}^{v_s}\}_{s=1}^{S}$ as key and values. This allows the model to produce a weighted visual summary $\tilde{\mathbf{Z}}^{v}$ that emphasizes the most relevant speaker-specific visual cues. The fused representation $\mathbf{Z} = \mathrm{Concat}(\mathbf{Z}^{a}, \tilde{\mathbf{Z}}^{v})$ is then fed into the audio-visual ASR model. Subsequent work \cite{2022avasrWsd} extends this by adding an auxiliary active-speaker detection task, using the same fused representation $\mathbf{Z}$ to jointly predict the speaking face.
\\
\indent For multi-speaker ASR that distinguishes speakers in a sentence, where multiple speakers must be differentiated within a single mixture, both SIMO and SISO systems benefit from visual conditioning, although in different ways. In the SIMO architecture, \cite{wu21e_interspeech} injects visual representations $\mathbf{Z}^{v}$ into both the encoder and decoder using the same cross-modal attention module. Their experiments show that it significantly outperforms simply concatenating all face-track features before the mixture encoder $\mathrm{Encoder}_{\text{Mix}}$. Visual information also helps resolve output-stream permutation. In SISO settings, where all speakers are decoded in a single transcript, \cite{makishima25b_interspeech} concatenates all available visual features with the audio representation prior to decoding, similar to single-speaker audio-visual ASR, and reports consistent performance gains.\\
\subsection{LLM-based text conditioning}
Recently, the emergence of multi-modal large language models (MLLMs)~\cite{large-speech-review} has provided a unified framework for a wide range of speech and language tasks. These models typically consist of a speech encoder and a projector that projects acoustic features into the LLM embedding space, followed by an LLM that performs sequence generation. Like single-speaker ASR with multimodal LLM~\cite{trinh2024discretemultimodaltransformerspretrained}, multi-speaker ASR can also be cast into this unified architecture~\cite{li_2023z--adapting_2023, shi2025serializedoutputpromptinglarge}. Because the unified output format of SISO models naturally aligns with the generation paradigm of MLLMs, current LLM-based multi-speaker systems almost exclusively adopt the SISO output format.\\
\indent MLLM-based systems benefit from instruction-based multi-task learning, which enhances speech understanding and enables flexible, user-defined interaction scenarios. The recent work of Meng et al.~\cite{meng_2024--large_2024} exemplifies this interactive paradigm: users can specify multi-speaker ASR tasks through natural language instructions, such as transcribing only the first speaker, a female speaker, or utterances containing specific keywords. The system integrates speech representations from the Whisper encoder and WavLM with an LLM. Compared to training with only a generic multi-speaker ASR instruction, this instruction-based training with additional contextual prompts improves the model's speech understanding and overall multi-speaker ASR performance.\\
\indent In addition, MLLMs can ingest auxiliary textual context that benefits both speech and speaker recognition. Long-tail or rare words can be supplied as external context; prior work in single-speaker ASR~\cite{trinh2025improvingnamedentitytranscription} demonstrates the effectiveness of providing such terms to a multimodal LLM, and~\cite{CMT_LLM2025} extends this idea to the multi-speaker setting with similar gains. Diarization information can also be integrated into an MLLM to assist multi-speaker decoding, as shown in~\cite{Lin2025DiarizationAwareMA}. In this framework, the LLM consumes not only projected speech tokens but also structured diarization triplets (speaker embeddings, start indices, and end indices), each aligned to the LLM embedding space through learnable linear layers.

\section{Long-Form Multispeaker ASR} \label{long-form}

The previous section focuses on pre-segmented audio. We turn to continuous long-form audio in this section. Unlike pre-segmented data, long-form audio must be partitioned before being processed by either SIMO or SISO ASR models, and the results must be integrated into a coherent global transcription with consistent speaker identities.
This section addresses these two key challenges. An ideal segmentation algorithm should be computationally efficient while preserving linguistic coherence. The segmentation methods are introduced in Section \ref{long-form:segmentation methods}.
Section \ref{long-form:hypothesis stitching} discusses how to merge local hypotheses into a unified global transcription, with consistent speaker identities across all segments.

\subsection{Segmentation Methods} \label{long-form:segmentation methods}

Segmenting long-form multi-speaker audio is crucial for enabling accurate and efficient ASR. Strategies need to balance computational efficiency and preserve linguistic coherence. Existing methods typically rely on acoustic or semantic cues.

Voice activity detection (VAD) segments speech by detecting silent intervals based on acoustic features such as energy levels and spectral patterns. Its simplicity leads to widespread use in both cascade and end-to-end systems \cite{2021kanda__comparative_long_term, kanda_2021slt_investigation_2021, yu_2022itsp--comparative_2022}. However, VAD-based segments may still be excessively long, necessitating further clipping to meet the input requirements of downstream ASR models. Also, since VAD operates purely on acoustic cues, it may inadvertently split sentences at unnatural boundaries, particularly when speakers pause or hesitate mid-sentence, which can disrupt linguistic coherence and degrade the performance of subsequent ASR tasks.

Sliding window segmentation is another strategy, which processes audio using fixed-length windows and a pre-defined stride. Unlike VAD, the sliding window method generates uniformly sized segments ready for ASR models. However, it still risks disrupting semantic coherence by splitting sentences at arbitrary boundaries, a problem exacerbated by rigid window constraints. Additionally, shorter strides can inflate computational costs due to extensive overlap.

To address these limitations, recent studies incorporate semantic information into segmentation. For example, Cornell et al. \cite{cornell_2024icsp--one_2023} introduce an adaptive sliding window strategy, which inserts the special token \texttt{<trunc>} to mark truncation, and resumes from the last silence point to preserve linguistic continuity and reduce redundancy. Huang et al. \cite{Huang2022E2ESJ} predict segment boundaries in a streaming manner, based on both acoustic and text-level cues, enabling dynamic segmentation with minimal overhead.

\subsection{Hypothesis Stitching Methods} \label{long-form:hypothesis stitching}
Generating the final global hypothesis requires concatenating and aligning local transcriptions from segmented audio. A straightforward approach is to concatenate local transcriptions directly with VAD-segmented clips \cite{kanda_2022icsp--transcribe--diarize_2022,kanda_2021slt_investigation_2021,cornell_2024icsp--one_2023}.
When long audio is segmented with overlaps, the final global transcription cannot be obtained by simple concatenation due to redundancy and potential mismatch. 
High-confidence word selection can be employed in overlapping parts \cite{Chiu2019ACO}, while a neural-network-based hypothesis stitcher \cite{chang_2021icsp--hypothesis_2021} fuses segment outputs into a coherent transcription without requiring alignment.

Maintaining globally consistent speaker labels is essential for long-form multi-speaker ASR. Existing approaches rely on speaker profiles or embeddings learned jointly within the ASR model.
In profile-based methods, speaker identities can be resolved during speaker-attributed decoding when speaker enrollment is available. When enrollment is not possible, speaker profiles can be approximated through unsupervised clustering of embeddings from a pretrained speaker encoder \cite{kanda_2022icsp--transcribe--diarize_2022}. Even supplying dummy profiles, which do not appear in the input audio, can improve performance \cite{kanda_2021slt_investigation_2021}.
In the joint modeling approach, the multi-speaker ASR system outputs both transcriptions and speaker embeddings, typically by a multi-task learning framework. Global labels can be obtained by clustering the embeddings. In \cite{cornell_2024icsp--one_2023}, each window of the E2E DAST model provides local speaker transcription and diarization with time-averaged speaker embeddings. Meanwhile, \cite{Mao2020SpeechRA} demonstrates the advantages of jointly learning speaker embeddings and transcriptions for hour-long multi-speaker podcasts, using lexical cues to improve speaker label assignment. After processing all windows, final diarization and speaker embeddings can be obtained by clustering these time-averaged embeddings. This method eliminates the need for additional speaker embedding models and improves computational efficiency, but its performance is heavily dependent on embedding quality.

\section{Evaluation} \label{sec:evaluation}

Multi-speaker ASR research relies on both real-world and simulated datasets.
Real-world datasets are collected from natural conversation scenarios such as meetings or phone calls; they provide authentic acoustic conditions, spontaneous speech patterns, and natural overlaps. However, they are often small, noisy and domain-specific, posing challenges for early-stage model training and broader applicability. 
Simulated datasets are created by overlapping single-speaker recordings and introducing noise at the configurable overlapping rate and noise level, enabling scalable as well as controlled training and evaluation of overlap and noise robustness. To improve realism, Yang et al. \cite{yang_2023icsp--simulating_2023} leverages statistical language models to guide overlap patterns. Additionally, Moriya et al. \cite{moriya_2024l--alignment-free_2024} introduce on-the-fly data generation to support dynamic parameter adjustment and improve memory efficiency. 
Table \ref{tab:datasets} highlights commonly used multi-speaker ASR datasets.

\begin{table}[ht]
\centering

\resizebox{0.95\linewidth}{!}{ 
\begin{tabular}{c|lcccc}
\toprule
 & \textbf{Dataset} & \textbf{Scenario} & \textbf{Hours} & \textbf{Language} & \textbf{\# Speakers} \\
\midrule
\multirow{4}{*}{\textit{Real}} 
  & AMI & Meetings & 100 & English & 3--5 \\
  & AliMeeting~\cite{Alimeeting} & Meetings & 120 & Chinese & 2--4 \\
  & CallHome & Phone calls & 60 & Multilingual & 2 \\
  & LibriCSS~\cite{libricss} & Meetings & 10 & English & 8 \\
\midrule
\multirow{4}{*}{\textit{Sim}} 
  & WSJ0-2mix~\cite{DeepClustering} & Read speech  & 45 & English & 2 \\
  & LibriMix~\cite{librimix} & Read speech & 500 & English & 2--3 \\
  & LibriHeavyMix~\cite{jin_2024--libriheavymix_2024} & Read speech  & 20,000 & English & 2--4 \\
\bottomrule
\end{tabular}
}
\caption{\textit{Common real-world and simulated datasets. 
\textbf{\# Speakers} refers to the number of speakers per recording in real datasets, and the exact mixed speaker number per sample in simulated datasets. }}
\label{tab:datasets}
\end{table}

Evaluating multi-speaker ASR systems requires measuring both transcription accuracy and the system's ability to assign speaker labels.
\textbf{Word error rate (WER)} evaluates overall transcription accuracy by measuring insertions, deletions, and substitutions. \textbf{Character Error Rate (CER)} is adopted for languages with non-alphabetic scripts, while \textbf{Sentence Error Rate (SER)} captures the proportion of sentences containing any error. In multi-speaker settings, these metrics can be extended with special tokens for speaker changes, overlapping speech, timestamps, and speaker identities. This extension enables evaluation not only for content accuracy but also of the system's capability to handle speaker turns and overlaps.

In addition to standard WER, several evaluation metrics have been proposed to account for speaker distinctions in multi-speaker ASR. \textbf{Concatenated minimum-permutation WER (cpWER)} \cite{CHIME6-cpwer} concatenates utterances per speaker and computes the WER across all permutations of hypotheses and references, selecting the lowest:
\begin{equation}
\text{cpWER}
=
\min_{\pi \in \mathcal{P}(S)}
\frac{
\sum_{s=1}^{S} \mathrm{WER}(H^s, R^{\pi(s)}) \cdot N_{\pi(s)}
}{
\sum_{s=1}^{S} N_s
}
\end{equation}
where $\mathcal{P}(S)$ denotes all permutations of $\{1, \ldots, S\}$, $N_s$ is the number of words in the reference transcript $R^s$, and \(H^s\), \(R^{\pi(s)}\) denote the hypothesis and reference for speaker \(s\), respectively. cpWER jointly reflects transcription and speaker assignment accuracy. 
\footnote{Note: a previous version of this paper showed an incorrect formula; this has now been fixed.}
\textbf{Speaker-attributed WER (SA-WER)} further enforces speaker correspondence by requiring that hypotheses be matched with the reference of the correct speaker label. It penalizes speaker assignment errors and provides a stricter evaluation of speaker-aware performance. Among these, WER, cpWER, and SA-WER are increasingly strict in evaluation criteria.
To incorporate temporal alignment, Neumann et al. \cite{neumann_meeteval_2023} propose \textbf{time-constrained cpWER (tcpWER)}, which restricts word matching to a fixed time window in addition to speaker permutation, which requires real or estimated token timestamps.

\subsection{Performance Comparison of E2E Approaches}
\begin{table*}[htbp]
\centering
\scriptsize
\resizebox{1\textwidth}{!}{
\begin{tabular}{
p{1cm}
p{2.8cm} 
>{\centering\arraybackslash}p{1.8cm} 
>{\centering\arraybackslash}p{1.5cm} 
>{\centering\arraybackslash}p{1.1cm} 
>{\centering\arraybackslash}p{1.5cm} 
>{\centering\arraybackslash}p{1.5cm} 
>{\centering\arraybackslash}p{1cm} 
>{\centering\arraybackslash}p{1.1cm} 
>{\centering\arraybackslash}p{1cm} 
>{\centering\arraybackslash}p{1.1cm}
}
\toprule
\textbf{Dataset} &
\textbf{Model} &
\textbf{Input Gran.} &
\textbf{Framework} &
\textbf{Spk. Enrl.} &
\textbf{Mix Pre. Hrs} &
\textbf{Params Tr/To} &
\textbf{SDM dev} &
\textbf{SDM eval} &
\textbf{IHM dev} &
\textbf{IHM eval} \\
\toprule

\multirow{6}{*}{\textbf{AMI}}
&Conformer AED~\cite{kanda_2021itsp--large-scale_2021}& Utterance Group& SISO &\scalebox{0.75}{\usym{2613}}&900k& 50 / 50& 18.4 & 21.2 & 13.5 & 14.9 \\

&WavLM/wTSE\&JSM~\cite{huang_2023icsp--adapting_2023} & Utterance Group & Hybrid & $\checkmark$ & 58 & 13 / 108 & -- & -- & -- & 28.4\dag \\

&Adapted USM~\cite{li_2023z--adapting_2023} & Utterance Group & SISO & \scalebox{0.75}{\usym{2613}} & -- & 84 / 1630  & -- & 21.4 & -- & -- \\

&META-CAT~\cite{wang_2024--meta-cat_2024} & Utterance Group & SISO &  \scalebox{0.75}{\usym{2613}} & -- & 600 / 723 & -- & -- & -- & 22.8 \\
CMT-LLM~\cite{CMT_LLM2025} & Utterance Group & SISO & \scalebox{0.75}{\usym{2613}} & -- & 24 / 7300 & 30.0 & 32.9 & 21.9 & 22.8\\
&Transcribe-to-Diarize~\cite{kanda_2022icsp--transcribe--diarize_2022} & Long-Term &  SISO & \scalebox{0.75}{\usym{2613}} & -- & 146 / 146 & 22.6 & 24.9 & 15.9 & 16.4 \\

&SLIDAR~\cite{cornell_2024icsp--one_2023} & Long-Term & SISO & \scalebox{0.75}{\usym{2613}} & 5k & 655 / 655 & 21.8 & 24.5 & 14.2 & 15.6 \\

\toprule
\textbf{Dataset} &
\textbf{Model} &
\textbf{Input Gran.} &
\textbf{Framework} &
\textbf{Spk. Enrl.} &
\textbf{\# Tr. Spk} &
\textbf{Params Tr/To} &
\textbf{2spk dev} &
\textbf{2spk eval} &
\textbf{3spk dev} &
\textbf{3spk eval} \\
\toprule
\multirow{13}{*}{\makecell[c]{\textbf{Libri-}\\\textbf{speech-}\\\textbf{Mix}}}
&LSTM SOT~\cite{kanda_2020itsp--serialized_2020} & Utterance Group& SISO & \scalebox{0.75}{\usym{2613}} & 1, 2, 3& 136 / 136 & -- & 11.2* & -- & 24.0* \\

&Transformer SOT~\cite{2021itsp_kanda_e2eTransformer}& Utterance Group& SISO & \scalebox{0.75}{\usym{2613}} &1, 2, 3 &  129 / 129 & -- & 4.9* & -- & 6.2* \\

&LSTM SA-ASR~\cite{kanda_2020itsp--joint_2020} & Utterance Group & SISO & $\checkmark$ & 1, 2, 3& 146 / 146  &--&9.9\dag&--&23.1\dag\\

&SA-MBR~\cite{kanda_2021icsp--minimum_2021}& Utterance Group& SISO & $\checkmark$ & 1, 2, 3 & 146 / 146 &--&9.5\dag&--&20.7\dag \\

&Transformer SA-ASR~\cite{2021itsp_kanda_e2eTransformer}& Utterance Group& SISO &$\checkmark$ & 1, 2, 3&  142 / 142 & -- & 6.4\dag & -- & 8.5\dag \\

&W2V-Sidecar~\cite{meng_2023ICSP_sidecar} & Utterance Group& SIMO & \scalebox{0.75}{\usym{2613}} & 2  & 9 / 104 & 6.0 & 5.7 & -- & -- \\

&CSE Network~\cite{kang_2024icsp--cross-speaker_2024}& Utterance Group& SIMO & \scalebox{0.75}{\usym{2613}} & 1, 2 & 33 / 33 & 11.8 & 10.7 & 24.2 & 24.3 \\

&CIF SA-SOT~\cite{fan_2024icsp-sa-sot_2024}& Utterance Group & SISO & \scalebox{0.75}{\usym{2613}} &2& 136 / 136 & -- & 3.4 & -- & -- \\

&SA-CTC SOT~\cite{kang_2024--disentangling_2024}& Utterance Group & SISO & \scalebox{0.75}{\usym{2613}} & 2  & 56 / 56 & 3.9 & 4.1 & 22.6 & 22.6 \\

&SD-CTC SOT~\cite{sdctc} & Utterance Group& SISO & \scalebox{0.75}{\usym{2613}} & 1,2 & 114 / 114 & -- & 3.5 & -- & -- \\
&Whisper-SS-TTI~\cite{meng_2024--empowering_2024}& Utterance Group & SIMO & \scalebox{0.75}{\usym{2613}} & 2, 3& 9 / 250 & --& 5.2 & -- & 8.6 \\
&Whisper-SS-TTI~\cite{meng_2024--empowering_2024}& Utterance Group & SIMO & \scalebox{0.75}{\usym{2613}} & 2, 3 &  13 / 779 &--& 4.0 & -- & 7.5 \\
&Whisper-SS-TTI~\cite{meng_2024--empowering_2024}& Utterance Group& SIMO & \scalebox{0.75}{\usym{2613}} & 2, 3& 18 / 1950 & --& 3.4 & -- & 6.8 \\

&MT-LLM~\cite{meng_2024--large_2024} & Utterance Group& SISO & \scalebox{0.75}{\usym{2613}} &2, 3 &  76.6 / 7550 & -- & 5.2  & -- & 10.2 \\

\toprule
\textbf{Dataset} &
\textbf{Model} &
\textbf{Input Gran.} &
\textbf{Framework} &
\textbf{Spk. Enrl.} &
\textbf{\# Tr. Spk} &
\textbf{Params Tr/To}&
\textbf{2spk dev} &
\textbf{2spk eval} &
\textbf{3spk dev} &
\textbf{3spk eval} \\
\toprule
\multirow{8}{*}{\textbf{LibrMix}}
&WavLM/wTSE\&JSM~\cite{huang_2023icsp--adapting_2023} & Utterance Group & SIMO & $\checkmark$ &2 & 13 / 108 &-- & 10.7\dag &--&-- \\

&Whisper-SS-TTI~\cite{meng_2024--empowering_2024} & Utterance Group &SIMO & \scalebox{0.75}{\usym{2613}} & 2, 3 & 9 / 250 & --& 9.4 & -- & 26.8 \\
&Whisper-SS-TTI~\cite{meng_2024--empowering_2024}&Utterance Group & SIMO & \scalebox{0.75}{\usym{2613}} & 2, 3&  13 / 779 & -- & 6.6 & -- &21.5 \\
&Whisper-SS-TTI~\cite{meng_2024--empowering_2024}&Utterance Group & SIMO & \scalebox{0.75}{\usym{2613}} &2, 3 & 18 / 1950 & --& 4.7 & -- & 16.8 \\

&W2V-Sidecar~\cite{meng_2023ICSP_sidecar} &  Utterance Group& SIMO & \scalebox{0.75}{\usym{2613}} &2 & 9 / 104 & 7.7 & 8.1 & -- & --\\

&GEncSep~\cite{shi_2024--serialized_2024-1} & Utterance Group& SISO& \scalebox{0.75}{\usym{2613}} & 2, 3 & -- & 6.4 & 6.6 & 13.3 & 13.1  \\
&CMT-LLM~\cite{CMT_LLM2025} & Utterance Group & SISO &\scalebox{0.75}{\usym{2613}} & 2 & 24 / 7300 & 7.3 & 8.1 & --& --\\
&Hypothesis Stitcher~\cite{chang_2021icsp--hypothesis_2021} & Long-term & SISO & \scalebox{0.75}{\usym{2613}} & 1 - 6 & -- &  -- & 11.5 & -- & 13.4 \\

&Hypothesis Clustering~\cite{kashiwagi_2024--hypothesis_2024} & Long-term &SISO&$\checkmark$ &1, 2, 3 & -- & -- &8.2\dag&--& 21.5\dag \\

\bottomrule
\end{tabular}
}
\caption{\textit{cpWER of multi-speaker ASR models on \textbf{AMI}, \textbf{LibrispeechMix}, and \textbf{LibriMix} corpus.
\textbf{SDM} denotes single distant microphone; \textbf{IHM} denotes mixture of independent headset microphones. 
\textbf{Mix Pre. Hrs} indicates the hours of simulated mixture data used for pretraining.
\textbf{Params Tr/To} indicates trainable / total model parameters in millions.
\textbf{\# Tr. Spk} refers to the speaker configuration used during training.
Methods are grouped by input granularity and listed chronologically.
* and \dag\ signify the paper reported results as standard WER or SA-WER, respectively.
}}
\label{tab:Results}
\end{table*}

This section compares various techniques and their accuracy across standard multi-speaker benchmarks. We report results on both real-world data (AMI) and simulated mixtures (LibriSpeechMix and LibriMix), as shown in Table~\ref{tab:Results}. LibrispeechMix~\cite{kanda_2020itsp--serialized_2020} provides the standard dev/test set for evaluation. To facilitate comparison between methods, we adopt cpWER as the primary evaluation metric to include speaker determination in WER. Note that some methods only report results on standard WER and SA-WER (see the results marked with * and \dag); those accuracies cannot be directly compared with those of other methods. All results are reported directly from the original papers.
Our analysis examines how different scenarios and training approaches impact results. 

The results demonstrate no consistent superiority between SISO and SIMO approaches in multi-speaker ASR. For instance, SISO methods \cite{kanda_2021itsp--large-scale_2021, shi_2024--serialized_2024-1} outperform SIMO \cite{huang_2023icsp--adapting_2023, meng_2023ICSP_sidecar} on AMI and LibriMix respectively, while SIMO method \cite{meng_2024--empowering_2024} surpasses SISO \cite{2021itsp_kanda_e2eTransformer} on LibriSeechMix. Moreover, cpWER has not shown consistent improvement throughout the six-year development period of end-to-end multi-speaker ASR approaches. The currently best performance on AMI comes from a relatively small 50M model trained on extensive 900k hours of simulated mixture data \cite{kanda_2021itsp--large-scale_2021} in 2021. This suggests a stagnation in real-world benchmark progress, despite many recent methodological proposals.

In general, contemporary research on multi-speaker ASR is less about pursuing marginal WER improvements on specific benchmark datasets. Instead, it tends to follow three trends: 
(1) {\bf Understanding scenario-dependent factors}: For example, LibriMix methods demonstrate this evolution: while early works \cite{kanda_2020itsp--serialized_2020} used no speaker enrollment, subsequent studies \cite{kanda_2020itsp--joint_2020, 2021itsp_kanda_e2eTransformer} introduced speaker-attributed (SA) methods with enrollment. This was further advanced by CIF SA-SOT \cite{fan_2024icsp-sa-sot_2024}, which achieved speaker attribution without enrollment. 
Meanwhile, Transcribe-to-Diarize \cite{kanda_2022icsp--transcribe--diarize_2022} extends the approach from \cite{kanda_2020itsp--joint_2020} to long-form scenarios. 
(2) {\bf Exploring novel architectures and information fusion methods}, such as cross-speaker encoding for SIMO \cite{kang_2024icsp--cross-speaker_2024}, enhancing CTC loss with speaker attribute \cite{kang_2024--disentangling_2024}.  
(3) {\bf Efficiently adapting single-speaker foundation ASR systems to multi-speaker settings with minimal training}, such as Adapted USM \cite{li_2023z--adapting_2023}, WavLM/wTSE\&JSM \cite{huang_2023icsp--adapting_2023}, Whisper-SS-TTI \cite{meng_2024--empowering_2024}, W2V-Sidecar \cite{meng_2023ICSP_sidecar}. The \# Parameters  (Train / Total) in tables partially reflect the reduced training effort, though they don't fully capture cases like META-CAT \cite{wang_2024--meta-cat_2024}, which achieves efficiency through fewer training epochs despite its larger trained size.

Currently, limited open-source availability makes fair comparisons difficult and slows research progress. Lots of methods show their improvements by comparing with their own baseline systems that don't include the new architectures. This highlights the importance of developing standardized benchmarks and sharing reproducible models as a community. A coordinated effort toward standardized open-source benchmark suites and consistent experiment protocols would substantially improve comparability and accelerate progress in multi-speaker ASR.

In addition to standardized benchmarks, several open-source toolkits provide practical starting points for building and evaluating multi-speaker ASR systems. 
ESPnet~\cite{watanabe2018espnet}, SpeechBrain~\cite{speechbrain}, NVIDIA NeMo~\cite{kuchaiev2019nemo}, and Pyannote~\cite{pyannote2020} respectively provide SOT recipes, audio-visual ASR pipelines, integrated diarization–ASR modules, and state-of-the-art diarization components. These resources further support reproducibility and practical usage.

\section{Conclusion and Future Directions}
Recent research on multi-speaker ASR has increasingly focused on end-to-end (E2E) methods, aiming to produce more accurate speaker-distinguishable transcriptions in scenarios such as phone calls, meetings, and group discussions. E2E methods can overcome the limitations of traditional modular systems, such as error propagation and failure to leverage cross-task synergies. Furthermore, progress in single-speaker ASR, speech separation, and diarization has provided the architectural foundation and pretrained model for initialization that facilitate the training of a multi-speaker ASR system, although with the scarcity of large-scale multi-speaker data.

This review provided an overview of end-to-end multi-speaker ASR approaches, from segment-level methods to processing continuous long-form recordings. Various architectural designs (Section~\ref{sec:Multi-speaker ASR}) were described for how to disentangle different speakers, and how to effectively leverage contextual cues from mixed speech, such as inter-speaker, overlapping, and temporal dependencies.
Beyond acoustic-only designs, Section~\ref{sec:multimodal} surveyed emerging multi-modal extensions, ranging from audio-visual fusion to LLM-based text conditioning, which provide complementary cues for speaker attribution and linguistic disambiguation, especially under heavy overlap or noise.
Recent work on long-form processing (Section~\ref{long-form}), which has further improved applicability to real-world cases, was also presented. 
Our accuracy comparisons indicate that, while end-to-end models often outperform modular approaches, no single architecture consistently outperforms others across end-to-end designs. Ongoing research continues to explore more sophisticated designs leveraging complex interaction patterns, while integrating advances in large-scale pretrained ASR models. Also, practical deployment factors, such as latency, streaming constraints, and computational cost, are becoming increasingly important and warrant further attention.

Overall, the future of the multi-speaker ASR lies in developing more robust, adaptive, and scalable systems for various scenarios. While recent designs have made significant progress, key challenges remain in robustly handling overlapping speech, and designing effective SISO/SIMO hybrids (Section~\ref{subsec:Future}). Moreover, future research may also explore adaptive modeling, such as transforming single-speaker models into multi-speaker systems without performance degradation, and optional speaker enrollment. In addition, advancing multi-speaker understanding -- through joint training with downstream objectives, cross-modal fusion (Section~\ref{sec:multimodal}), and injecting multi-speaker ASR into a unified multi-task foundation model.

{\bf Acknowledgment}:
This research was supported by the NSF National AI Institute for Student-AI Teaming (iSAT) under grant DRL \#2019805, and also from grant \#2046505. The opinions expressed are those of the authors and do not represent the views of the NSF. Finally, we are grateful for the feedback from the anonymous reviewers.

\bibliography{reference.bib}{}

\end{document}